# Conservative AI for Safety-Sensitive Medical Image Restoration: Residual-Bounded CT/CTA Enhancement for Intracranial Aneurysm–Relevant Signal Recovery

Weijun Ma
Independent Researcher
King George, Vancouver School Board
Vancouver, BC, Canada
Corresponding Author: weijun.ma@ieee.org

## Abstract

Image restoration models are increasingly applied to degraded medical scans, but in safety-sensitive contexts, these models are expected not only to improve image-quality metrics but also avoid uncontrolled modification of clinically important regions. This is especially relevant for intracranial computed tomography (CT) and CT angiography (CTA), where small vessels and aneurysm-relevant cues lie close to high-contrast bone and parenchymal boundaries, and where over-correction can plausibly remove or distort the very signal that human readers rely on. We frame medical image restoration as a conservative AI problem and present a residual-bounded 2.5D restoration framework trained on synthetically degraded CT/CTA inputs. The model adds a learned residual to the original center slice through an edit-control map that limits the magnitude and spatial extent of modification, which is intended to constrain restoration behavior rather than to produce free image generation. We evaluate the framework with a multi-level protocol that combines an aneurysm-relevant image-recovery matrix, paired comparison against a Gaussian baseline, Monte Carlo stability testing under repeated stochastic degradation, anatomical localization of meaningful edits, and external evaluation on low-dose CT. On 50 out-of-distribution CT/CTA cases the bounded model achieved a mean target gain of 0.0635, a mean PSNR of 37.51 dB, and an iatrogenic-edit rate of 4.0%; over 1,000 Monte Carlo runs (100 cases × 10 seeds) it remained net positive in 85.4% of runs with no stably negative cases. On external low-dose CT the model was directionally beneficial and produced a substantially smaller modification footprint than the baseline. Meaningful edits concentrated in brain and skull regions while unrelated anatomy showed negligible change. These findings are preliminary computational evidence that residual-bounded, conservatively evaluated restoration is feasible in boundary-sensitive vascular imaging; they do not establish clinical diagnostic performance and require expert review and prospective validation before any clinical use.

# 1. Introduction

Generative and restoration models are increasingly used to enhance degraded medical scans, but in safety-sensitive applications a restoration model is not free to rewrite its input. Clinically important cues are often subtle, and uncontrolled modifications near anatomical boundaries can introduce edits whose downstream effect on human interpretation is difficult to bound. This concern motivates a conservative AI perspective on medical image restoration: a useful model is one that improves image-level quality while constraining the magnitude, location, and stability of its own modifications. From this perspective, restoration is closer to controlled correction than to free image generation, and evaluation should explicitly probe how much, where, and how reliably the model edits.

Intracranial CT and CTA imaging is a natural test case for this perspective. CT and CTA are widely used to evaluate suspected intracranial aneurysms because of their availability, speed, and broad coverage of cerebrovascular anatomy (Thompson et al., 2015; Levinson et al., 2023). Real-world scans, however, are often affected by low-dose noise, blur, motion, ring or band artifacts, and reconstruction artifacts, which can attenuate the subtle vascular signals that small aneurysms produce on imaging (Din et al., 2023; Yang et al., 2021). Because aneurysm-relevant cues frequently sit on or near high-contrast bone–vessel and vessel–parenchyma boundaries, an aggressively enhancing model can plausibly improve global metrics while smoothing or shifting the very boundaries that are most informative for interpretation (Bizjak & Špiclin, 2023; Joo, 2025). This is the regime in which conservative behavior matters most.

In this work we treat low-quality CT/CTA restoration as a controlled-modification problem rather than a maximum-quality reconstruction problem. We train a 2.5D residual-bounded model on synthetically degraded CT/CTA inputs, where degradations are designed to imitate plausible CT failure modes rather than arbitrary corruptions, and we evaluate the model along several axes that together describe its restoration behavior: aneurysm-relevant image recovery, paired comparison against a conventional Gaussian baseline, stability under repeated stochastic degradation, the spatial concentration of meaningful edits, and external behavior on low-dose CT. We emphasize at the outset that the evaluation reported here is computational. It is intended to characterize controlled restoration behavior, not to establish clinical diagnostic performance.

**Contributions.** This paper makes the following contributions:

1. A residual-bounded conservative restoration design for low-quality CT/CTA, in which a learned residual is added to the original center slice through an edit-control map that limits both the magnitude and the spatial extent of modification.
2. A synthetic degradation pipeline for controlled evaluation, which generates paired degraded–clean inputs from relatively clean CT/CTA scans using blur, motion, Poisson–Gaussian noise, ring/band artifacts, and edge-streak artifacts.

3. A multi-level evaluation protocol, combining an aneurysm-relevant image-recovery matrix, paired comparison against a Gaussian baseline, Monte Carlo stability testing, anatomical-overlap localization of meaningful edits, and external low-dose CT evaluation.
4. A conservative interpretation of the results, in which we treat modification footprint, stability, and anatomical localization as primary evidence about restoration behavior, and treat PSNR as one image-level proxy among several rather than as a stand-alone endpoint.

A shorter version of this work has been accepted for presentation at the 2026 3rd International Conference on Artificial Intelligence and Future Education (AIFE 2026), Tokyo, Japan. The present manuscript reports an expanded computational imaging preprint version of the work.

# 2. Related Work

## 2.1 AI for CT/CTA aneurysm imaging

A growing body of work applies deep learning to intracranial aneurysm detection and segmentation on CT and CTA. Systematic reviews and meta-analyses have summarized the rapid expansion of this literature (Bizjak & Špiclin, 2023; Din et al., 2023; Joo, 2025). Reported systems range from research prototypes evaluated on curated datasets (Yang et al., 2021) to commercial-grade tools assessed in stepwise multicenter or workflow-oriented studies (Heit et al., 2022; Hu et al., 2024). Most of this work focuses on detection or segmentation performance rather than on the behavior of the restoration or preprocessing stage that may precede such systems, and the sensitivity of downstream models to input quality is often discussed only at a high level.

## 2.2 Deep learning restoration for low-dose or degraded CT

A parallel literature addresses denoising, super-resolution, and reconstruction for low-dose or degraded CT. Convolutional and encoder–decoder networks were among the early supervised approaches (Chen et al., 2017), and subsequent work has extended these methods to low-dose-specific settings, model-based reconstruction, and hybrid pipelines (Chen, Li, Zhou, & Li, 2024; Koetzier et al., 2023; Kulathilake et al., 2023). These methods typically report PSNR, SSIM, or other image-quality metrics, and many achieve substantial improvements relative to degraded inputs. Two practical issues recur. First, true paired degraded–clean clinical data are rarely available at scale, which complicates direct supervised training. Second, image-quality metrics do not, by themselves, characterize the behavior of the restoration model in regions where small modifications can plausibly affect interpretation, such as the immediate neighborhood of vessels and aneurysm candidates.

## 2.3 Conservative AI and safety-sensitive restoration evaluation

Existing CT/CTA restoration evaluations rarely combine image-recovery metrics with explicit measures of modification footprint, robustness under stochastic degradation, and anatomical localization of edits. The present work is positioned to complement that literature rather than to claim novelty in detection or reconstruction. We focus on controlled restoration behavior in a boundary-sensitive vascular imaging context: how strongly the model modifies its input, how stable that behavior is under repeated random degradation, where in the image meaningful edits concentrate, and whether the same restoration profile is preserved when the model is exposed to external low-dose data drawn from a different distribution. Treating these properties as first-class evaluation targets is consistent with a conservative AI view in which the safety of a restoration model is partly a function of how restrained and how predictable its modifications are.

# 3. Methods

This section describes the data sources used in this study, the synthetic degradation pipeline, the residual-bounded 2.5D restoration model, the training objective, and the multi-level evaluation protocol. The goal is a transparent description that supports reproducibility rather than a claim of clinical readiness.

## 3.1 Datasets and Evaluation Roles

Four data sources played complementary roles in model development and evaluation. The development pool consisted of CT/CTA scans drawn from a public intracranial aneurysm imaging resource and used both for restoration training and for in-distribution image-recovery evaluation. External low-dose CT data were used to probe behavior outside the synthetic training setting. Anatomical segmentation masks produced by an off-the-shelf segmentation tool were used in a post hoc analysis to ask whether meaningful edits concentrated in plausible anatomical regions rather than spreading across the image. Synthetically degraded scans, generated from relatively clean inputs, supplied paired degraded–clean training data. Table 1 summarizes these data sources and their roles.

*Table 1. Datasets and their roles in model development and evaluation.*

| Dataset | Data Type | Role in Study | Evaluation Stage | Main Purpose |
|---|---|---|---|---|
| RSNA Intracranial Aneurysm Detection | CT/CTA | Main dataset for model development and image-recovery evaluation | Training / in-distribution evaluation | Train the restoration model and assess aneurysm-relevant signal recovery |
| Mayo low-dose CT dataset | Low-dose CT | External dataset for generalization testing | External testing | Evaluate transfer beyond the synthetic training setting |

| Dataset | Data Type | Role in Study | Evaluation Stage | Main Purpose |
|---|---|---|---|---|
| TotalSegmentator-based anatomical masks | Anatomical segmentation masks | Anatomical localization analysis | Post hoc evaluation | Assess whether meaningful edits concentrate in plausible anatomical regions |
| Synthetic degraded scans generated from relatively clean CT/CTA | Simulated degraded CT/CTA | Paired training input creation | Training / robustness testing | Create realistic degraded–clean pairs when true paired clinical data are unavailable |

## 3.2 Ethics and Data Governance

This study used publicly available and/or de-identified imaging datasets. No new patient recruitment, patient contact, clinical intervention, or prospective collection of human data was carried out. No directly identifying patient information was used, displayed, or released. Dataset use followed the terms of the original data providers. The model and its outputs are intended for research use only and are not used for clinical diagnosis, treatment planning, triage, or radiology workflow deployment.

## 3.3 Synthetic Degradation Generation

Because true paired degraded–clean clinical CT/CTA data are rarely available, paired training inputs were created by applying synthetic degradations to relatively clean CT/CTA scans. The degradation pipeline was designed to imitate failure modes that recur in CT imaging rather than arbitrary corruptions, and combined Gaussian and motion blur, Poisson–Gaussian noise, ring or band artifacts, and edge-streak artifacts. This design retains a meaningful restoration target while keeping the input distribution within a controlled and traceable family of perturbations. We do not claim that synthetic degradation reproduces the full distribution of clinical artifacts, and this limitation is discussed explicitly in Section 6.

## 3.4 Conservative 2.5D Restoration Model

The restoration model takes a 2.5D input consisting of the previous, center, and next axial slices and produces a restored center slice. This design provides local anatomical context without requiring a full 3D volumetric pass, and is consistent with a conservative rather than generative posture: the model is asked to correct a single slice in place, using its neighbors as context. Figure 1 summarizes the overall pipeline, which links input degradation, 2.5D conservative restoration, and the four-level evaluation protocol described in Section 3.7.

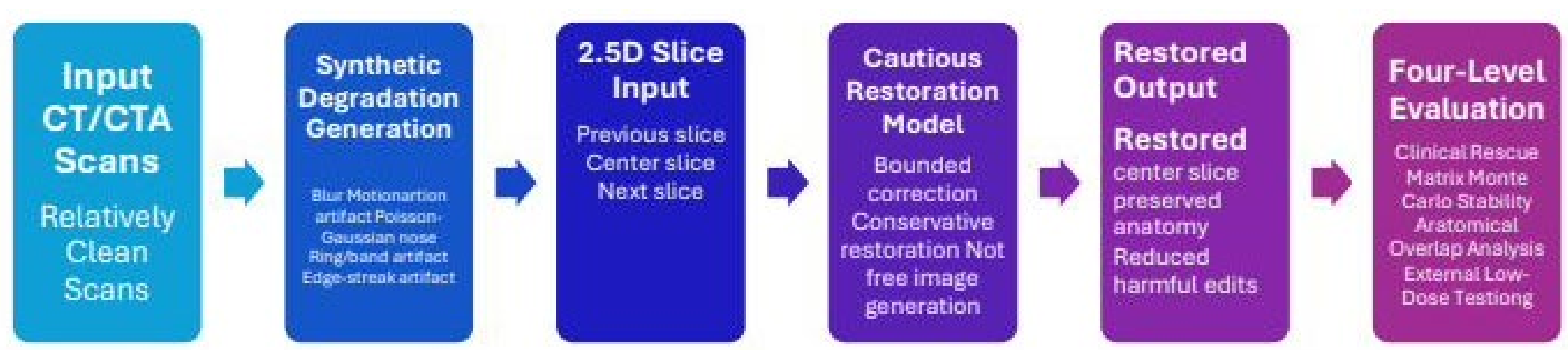

*Figure 1. Overall study pipeline. Relatively clean CT/CTA scans are passed through a synthetic degradation generator (blur, motion, Poisson–Gaussian noise, ring/band, edge-streak) to produce paired inputs. A 2.5D residual-bounded restoration model corrects the center slice using the previous, center, and next slices as context. Restored outputs are evaluated under four complementary analyses: aneurysm-relevant image-recovery matrix, Monte Carlo stability, anatomical-overlap localization, and external low-dose CT testing.*

## 3.5 Residual-Bounded Correction

Rather than producing a restored image directly, the model emits a learned residual that is added to the original center slice through a learned edit-control map. The map is intended to limit both how large and where modifications occur. Concretely, the restored center slice is given by Equation (1):

$$\hat{y} = x_c + m \odot r$$

*(1)*

where $\hat{y}$ is the restored center slice, x_c is the original (degraded) center slice, r is the learned residual correction, m is the edit-control map that limits the strength and location of modification, and $\odot$ denotes element-wise multiplication. Because the output is constructed by adding a bounded residual to the original input rather than by regenerating the slice from scratch, the model is structurally biased toward leaving most of the image unchanged. This is the property we associate with conservative restoration in a safety-sensitive setting.

## 3.6 Training Objective

The training objective combines a recovery term with explicit penalties on unnecessary or unstable modification, reflecting the principle that a useful restoration model in a safety-sensitive setting should not only improve image-level quality but also remain restrained where over-editing is plausibly harmful. Equation (2) defines the total loss:

$$L_total = w_1 \, L_restore + w_2 \, L_identity + w_3 \, L_edit + w_4 \, L_smooth + w_5 \, L_uncertainty$$

*(2)*

where L_restore encourages accurate recovery toward the clean target, L_identity penalizes unnecessary modification when the input is already close to clean, L_edit penalizes excessive correction magnitude, L_smooth regularizes the spatial behavior of the edit-control map, and L_uncertainty encourages caution in regions where the model is less confident. The weights $w_1$ through $w_5$ control the relative contribution of each term.

### 3.7 Evaluation Protocol

The model was evaluated among four complementary axes. First, an aneurysm-relevant image-recovery matrix (formerly referred to as a Clinical Rescue Matrix, hereafter the image-recovery matrix) was used to assess whether restoration improved aneurysm-relevant image-level signal relative to degraded input and a Gaussian baseline. Second, a Monte Carlo stability test repeatedly applied stochastic degradations to a fixed pool of cases and asked whether the restoration remained net positive across seeds, rather than depending on a single favorable corruption pattern. Third, an anatomical-overlap analysis used segmentation masks from an off-the-shelf anatomical segmenter to ask whether meaningful edits concentrated in plausible anatomical regions. Fourth, external low-dose CT testing asked whether the same restoration profile carried over beyond the synthetic training distribution. Together, these analyses are intended to characterize effectiveness, robustness, spatial selectivity, and external behavior, rather than to establish clinical diagnostic performance.

I must emphasize that the metrics used in this protocol are image-level proxies. They are informative about restoration behavior but do not, on their own, characterize aneurysm detection sensitivity or specificity in human readers.

## 4. Results

We report results along the four axes of Section 3.7: overall image-recovery performance, paired comparison against a Gaussian baseline, robustness under stochastic degradation, and external behavior together with anatomical localization. Numerical values are taken directly from the analyses run on the data described in Section 3.1.

### 4.1 Overview of Quantitative Findings

Table 2 summarizes the main quantitative findings across image-recovery evaluation, baseline comparison, robustness testing, anatomical localization, and external generalization. With the comparison show, the bounded model produced the strongest aneurysm-relevant image recovery profile among the compared settings, with the gains supported across stability, localization, and external evaluations. These observations are consistent with a model that improves image level signal while keeping its modification footprint small. The table here shows both results in numerical values, as well as interpretation of the result, more specifically as a comparison between the bounded model and the compared experiment baseline.

*Table 2. Main quantitative results across image-recovery evaluation, paired baseline comparison, robustness testing, anatomical localization, and external generalization.*

| Experiment | Key Metrics | Main Result | Interpretation | Evidence Type |
|---|---|---|---|---|
| Image-recovery matrix (N = 50 OOD CT/CTA cases) | Mean target gain, PSNR, iatrogenic-edit rate | Mean target gain = 0.0635; PSNR = 37.51 dB; iatrogenic-edit rate = 4.0% | The bounded model improved aneurysm-relevant signal while keeping modifications within a relatively low envelope. | Primary image-recovery evidence |
| Paired comparison vs. Gaussian baseline | Target-gain win rate, Δ target gain, Δ PSNR | Higher target gain in 64% of cases; Δ target gain = +0.0150; Δ PSNR = +1.856 dB | The bounded model outperformed a conventional baseline in most paired comparisons. | Baseline comparison evidence |
| Monte Carlo stability (100 cases × 10 seeds = 1,000 runs) | Positive-run rate, stably negative cases | Positive in 85.4% of runs; 0 stably negative cases | Performance remained net positive under repeated stochastic degradation. | Robustness evidence |
| Anatomical overlap analysis | Regions of meaningful edit concentration | Meaningful edits concentrated in brain and skull regions | Behavior consistent with targeted restoration rather than indiscriminate global smoothing. | Anatomical localization evidence |
| External low-dose testing | Transfer behavior on external data | Directionally beneficial transfer with smaller modification footprint than the baseline | Restoration profile preserved beyond the synthetic training distribution. | External generalization evidence |

## 4.2 Aneurysm-Relevant Image Recovery

Figure 2 reports the image-recovery matrix and paired comparison results for degraded input, a Gaussian baseline, and the bounded model. On 50 out-of-distribution CT/CTA cases the bounded model achieved a mean target gain of 0.0635, the highest mean PSNR among the compared settings at 37.51 dB, and an iatrogenic-edit rate of 4.0%. In paired comparisons against the Gaussian baseline, the bounded model produced a higher target gain in 64% of cases, with a mean improvement in target gain of +0.0150 and a mean improvement in PSNR of +1.856 dB. These results suggest that the bounded model recovers aneurysm-relevant image-level signal more effectively than the baseline while remaining within a relatively low iatrogenic-edit envelope. They should be interpreted as image-level proxy evidence rather than as a measure of diagnostic accuracy.

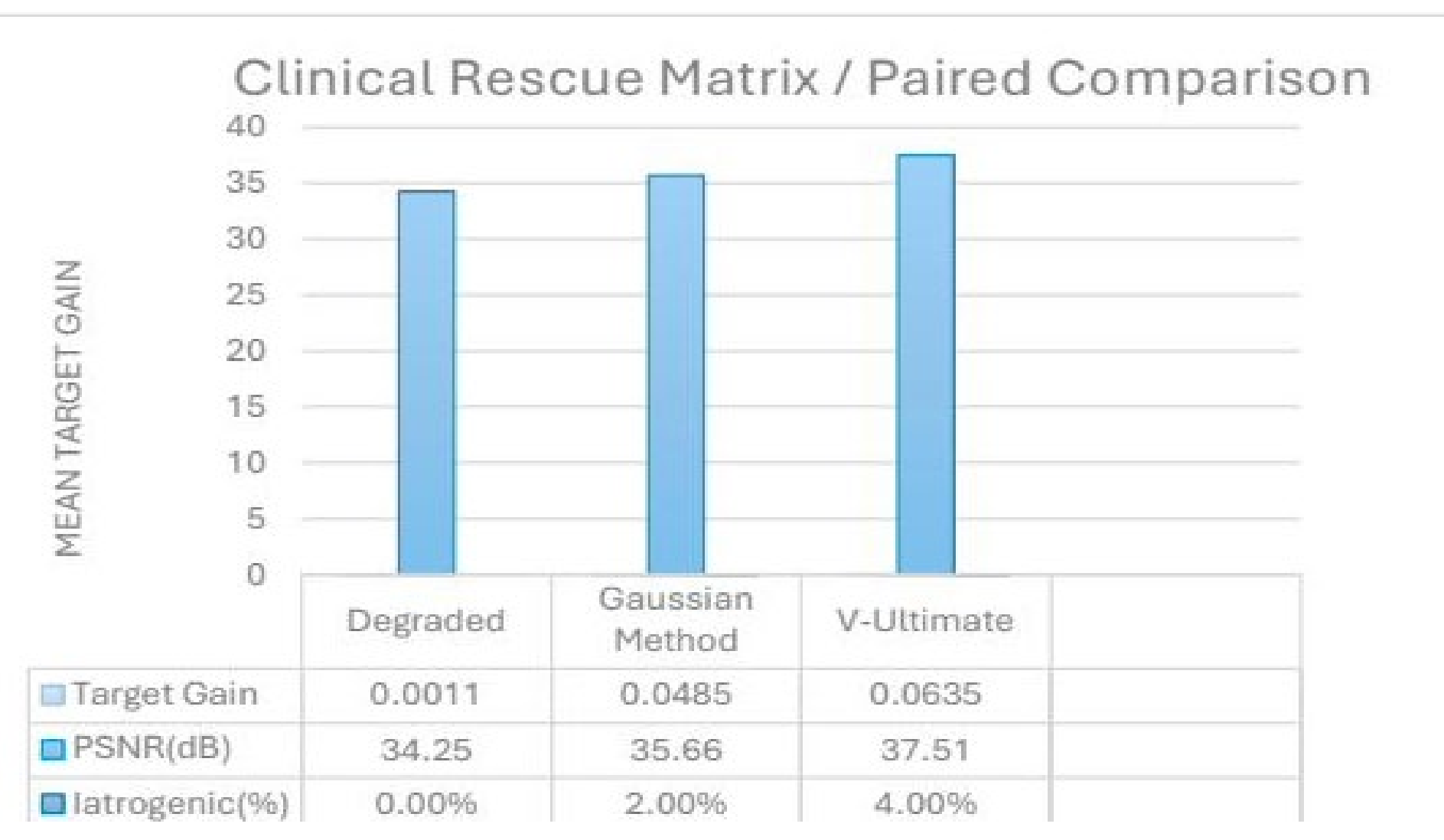

*Figure 2. Aneurysm-relevant image-recovery matrix and paired comparison. Mean target gain, PSNR (dB), and iatrogenic-edit rate for degraded input, a Gaussian baseline, and the bounded model on 50 out-of-distribution CT/CTA cases.*

## 4.3 Robustness Under Repeated Stochastic Degradation

Figure 3 summarizes the Monte Carlo stability analysis. Across 100 cases and 10 random seeds per case (1,000 runs in total), the bounded model was net positive in 85.4% of runs. At the case level, 54 cases were classified as stably positive, 45 as noise-sensitive, 1 as neutral, and 0 as stably negative. The absence of stably negative cases is informative: it indicates that the observed gains were not driven by a single favorable corruption pattern, and that performance did not collapse under any of the sampled degradation realizations. Noise-sensitive cases remained, however, which we interpret as a reminder that some inputs are intrinsically more variable under stochastic degradation and should not be presented as reliably restored.

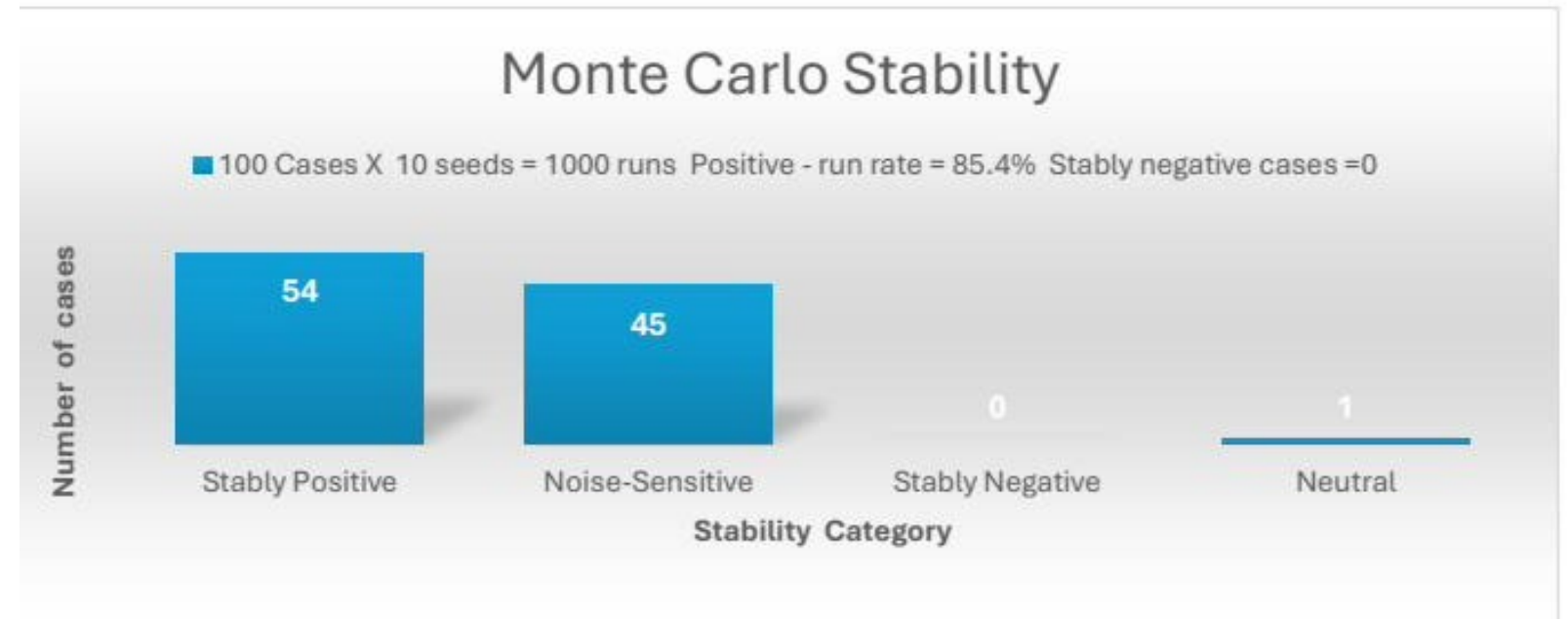

*Figure 3. Monte Carlo stability across 100 cases × 10 seeds (1,000 runs). Case-level distribution of stably positive, noise-sensitive, stably negative, and neutral outcomes; the run-level positive rate is 85.4% and there are no stably negative cases.*

## 4.4 External Low-Dose Behavior and Anatomical Localization

Table 3 summarizes the external low-dose CT evaluation and the anatomical-overlap analysis. On the external low-dose set, the bounded model improved PSNR over the quarter-dose input in paired comparison and was directionally beneficial relative to the Gaussian baseline, with a paired PSNR-improvement win rate of 87.5% versus 75.0% for the Gaussian baseline. A particularly relevant property in this context is the size of the modification footprint: the bounded model produced a maximum modification magnitude of 0.018, compared with 0.218 for the Gaussian baseline. A smaller modification footprint with comparable or better paired-comparison behavior is consistent with a more conservative restoration profile.

The anatomical-overlap analysis indicated that meaningful edits concentrated mainly in brain and skull regions, with mean segmentation-share values of 0.006646 and 0.006585 respectively, and with maximum segmentation shares of 0.019811 and 0.020330. Most unrelated anatomical regions showed negligible or zero meaningful change. This pattern is consistent with targeted restoration rather than indiscriminate global smoothing. We note that the segmentation masks used here are produced by an off-the-shelf anatomical segment and are not expert vascular annotations.

*Table 3. External low-dose CT testing and anatomical-overlap localization summary.*

| Evaluation Component | Metric | Main Result | Interpretation |
|---|---|---|---|
| External low-dose testing (Mayo dataset) | Mean PSNR | Quarter-dose input: 41.44 dB; Gaussian: 42.37 dB; bounded model: 41.84 dB; NLM: 17.24 dB | External performance remained competitive; the bounded model improved over degraded input while avoiding the instability seen in weaker baselines. |
| External paired evaluation | Paired PSNR improvement | Gaussian: +0.94 dB; bounded model: +0.40 dB | Both methods improved PSNR externally; the bounded model remained directionally beneficial under real low-dose conditions. |
| External paired evaluation | PSNR win rate | Bounded model: 87.5%; Gaussian: 75.0% | The bounded model showed stronger consistency across external paired comparisons. |

| Evaluation Component | Metric | Main Result | Interpretation |
|---|---|---|---|
| External modification footprint | Maximum modification magnitude | Gaussian: 0.218; bounded model: 0.018 | The bounded model achieved restoration with a substantially smaller modification footprint, consistent with a more conservative restoration profile. |
| Anatomical overlap analysis | Regions of meaningful edit concentration | Highest overlap observed in brain and skull regions | Behavior consistent with targeted restoration rather than indiscriminate global smoothing. |
| Anatomical overlap analysis | Mean segmentation share | Brain: 0.006646; skull: 0.006585 | Meaningful edits were concentrated in plausible anatomical regions while remaining small in scale. |
| Anatomical overlap analysis | Maximum segmentation share | Brain: 0.019811; skull: 0.020330 | Even the strongest localized edits remained spatially constrained. |
| Anatomical overlap analysis | Unrelated anatomical regions | Most unrelated structures showed negligible or zero meaningful change | The restoration process appeared spatially selective rather than globally aggressive. |

Figure 4 shows complementary representative qualitative restoration examples across blur levels and anatomical regions. Each row shows one case with the corresponding clean image, degraded input, Gaussian baseline, model output, and a region on interest for comparison between the baseline and the bounded model. Across these examples, the bounded model preserved sharper local boundaries and a more coherent local anatomy while avoiding a globally smoothed appearance, which was a common scene that a Gaussian baseline would produce. The region of interest panels (in red) highlights boundary sensitive regions that demonstrates significant differences, and it shows a clear comparison of the ability for the proposed method to restore scans in a clear and conservative method. It is important to note that these examples illustrate quantitative trends and cannot be used as a replacement or substitution for blinded reader evaluation. These comparisons are important to note, and shows key differences and improvements our proposed method provides, but real reader evaluation is required to demonstrate true differences for users of these models.

This figure aims to demonstrate examples that demonstrate the most varying set out anatomical regional data, it aims to compare examples from the most realistic sources possible, and is made in consideration of realistic locations of aneurysms to ensure the most realistic areas of necessary scans are shown.

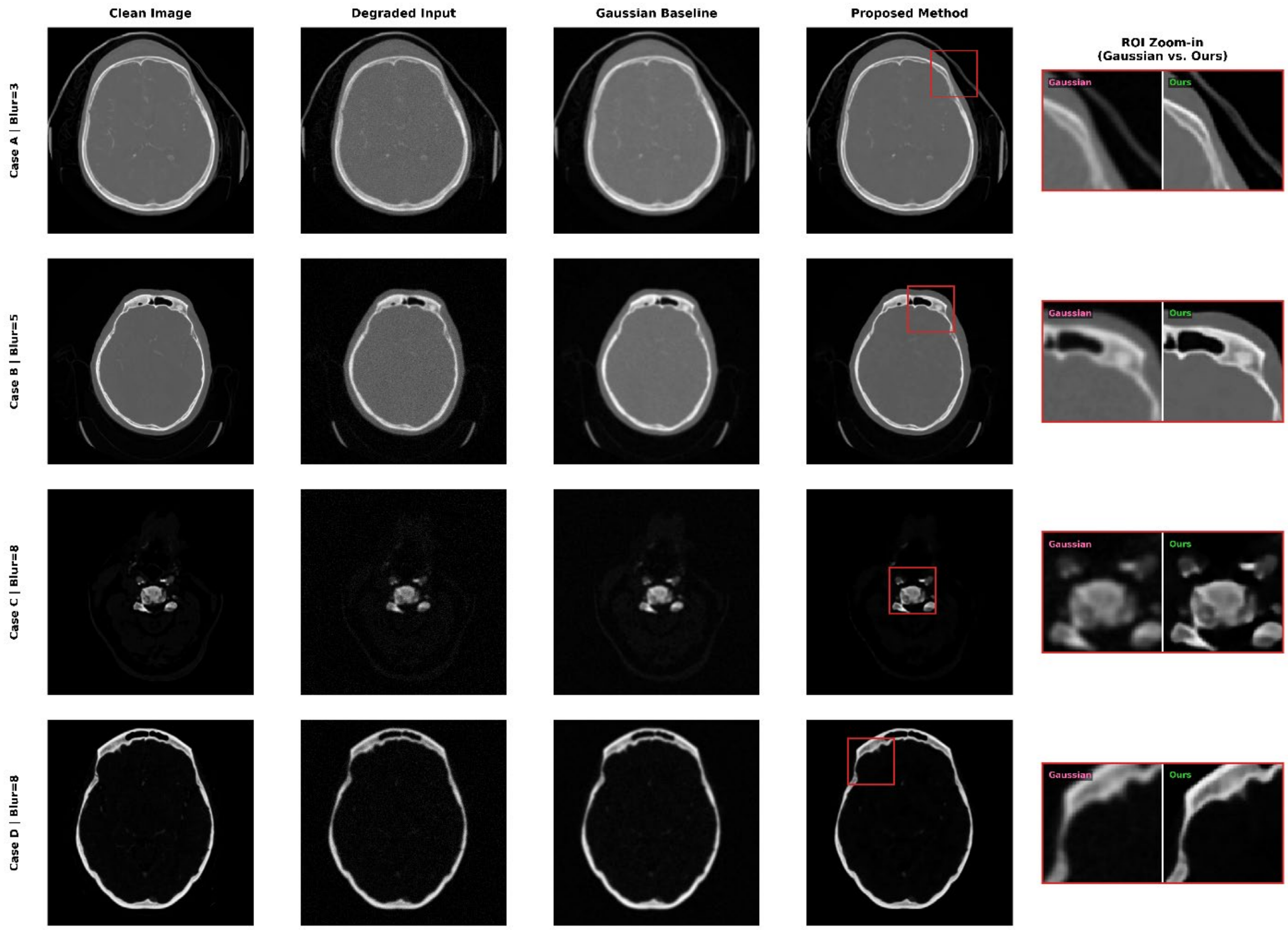

*Figure 4. Representative qualitative restoration examples. Each row shows one case (cases A–D, blur levels 3, 5, 8, 8) with the clean image, degraded input, Gaussian baseline, bounded model output, and a side-by-side region-of-interest zoom-in comparing the Gaussian baseline and the bounded model. Examples are illustrative of the quantitative trends and are not a substitute for blinded reader evaluation.*

# 5. Discussion

The results above describe a restoration model that improves aneurysm-relevant image-level signal, remains net positive under repeated stochastic degradation, concentrates meaningful edits in plausible anatomical regions, and carries a small modification footprint into an external low-dose setting. We discuss these findings under three headings.

**Conservative AI as controlled modification.** The framing we adopt treats restoration as a controlled-modification problem: a useful restoration model is one whose edits are bounded in magnitude, restricted in spatial extent, and stable under random variation in the input. The residual-bounded design and the composite training objective in Section 3 make this posture explicit. The empirical pattern in Section 4—improved target gain with a relatively low iatrogenic-edit rate, a 0/100 stably negative count, and a markedly smaller external modification footprint than the Gaussian baseline—is consistent with that posture. We do not claim that the model is intrinsically safe; we claim that, on the analyses reported here, its modification behavior is more restrained and more predictable than that of a representative non-bounded baseline.

**Limits of global image-quality metrics in boundary-sensitive imaging.** PSNR and similar global metrics are useful but insufficient in vascular CT/CTA. A model can in principle improve such metrics by smoothing or otherwise editing the immediate neighborhood of vessels and aneurysm candidates, which is precisely the region where small modifications are most likely to matter for human interpretation. For this reason we do not present PSNR as a stand-alone endpoint; rather, we treat it as one image-level proxy among several, alongside iatrogenic-edit rate, paired-comparison consistency, Monte Carlo stability, and anatomical localization. Reporting the modification footprint explicitly is, in our view, an important complement to image-quality numbers in this regime.

**Why robustness, modification footprint, and anatomical localization matter for safety-sensitive evaluation.** The contribution of this paper is not a claim that AI can diagnose intracranial aneurysms. The contribution is that AI restoration systems intended for safety-sensitive vascular imaging can be evaluated using a multi-evidence protocol—image recovery, paired baseline comparison, robustness, modification footprint, and anatomical localization—before any clinical claim is made. A restoration model that performs well on image-quality metrics but whose edits cannot be localized, whose behavior is unstable under repeated stochastic degradation, or whose modification footprint is large relative to a simple baseline, should not be considered ready for downstream clinical evaluation, regardless of its headline scores. The protocol presented here is one concrete instance of such an evaluation.

## 6. Limitations

This study has several limitations that should be considered when interpreting the results.

The training degradations are synthetically generated. They are designed to imitate plausible CT failure modes but do not reproduce the full distribution of real clinical artifacts, scanner-specific behavior, or patient-specific factors. Behavior on real clinical artifacts may therefore differ from what is reported here.

External low-dose CT testing is limited in scale. The results are directionally informative but do not constitute a large-scale external validation across multiple sites, scanners, or acquisition protocols.

We did not perform a blinded radiologist review. We did not perform direct evaluation of aneurysm detection sensitivity or specificity by human readers, and we make no claim about either.

The image-recovery matrix and related metrics are image-level proxies. They are informative about restoration behavior but do not characterize clinical diagnostic endpoints.

The anatomical masks used in the localization analysis come from an off-the-shelf anatomical segmenter rather than from expert vascular annotations. They support coarse localization claims (for example, that meaningful edits concentrate in brain and skull regions) but they do not provide vessel-level or aneurysm-level ground truth.

Taken together, these limitations imply that the present work should be interpreted as preliminary computational evidence about controlled restoration behavior, not as evidence of clinical readiness.

## 7. Conclusion

We presented a conservative residual-bounded AI framework for safety-sensitive medical image restoration and applied it to low-quality CT/CTA. Across an aneurysm-relevant image-recovery matrix, paired comparison against a Gaussian baseline, Monte Carlo stability testing, anatomical-overlap localization, and external low-dose CT evaluation, the bounded model improved image-level recovery while keeping its modification footprint small, remained net positive under repeated stochastic degradation, and concentrated meaningful edits in plausible anatomical regions. These findings support further investigation of controlled restoration in boundary-sensitive medical imaging. They do not establish clinical diagnostic performance. Future work should include larger-scale external validation, expert vascular annotations, blinded radiologist review, and direct evaluation of whether such restoration affects aneurysm detection in human readers to provide full context and genuine, realistic testing to show the ability of this proposed method in a conservative AI method of scan restoration.

## Code and Data Availability

Code, reproducibility materials, and project resources are publicly available through the project's GitHub repository (GitHub) and Zenodo archive (https://doi.org/10.5281/zenodo.19363880). These resources are intended to support transparency, reproducibility, and future external evaluation. The datasets used in this study are publicly available and/or de-identified, and were used in accordance with the terms of the original data providers.

## Ethics and Data Governance

This study used publicly available and/or de-identified imaging datasets. No new patient recruitment, patient contact, clinical intervention, or prospective collection of human data was carried out. No directly identifying patient information was used, displayed, or released. Dataset use followed the terms of the original data providers.

## Clinical Use Disclaimer

The model, code, figures, and results reported in this manuscript are for research purposes only. They are not intended for clinical diagnosis, treatment planning, triage, radiology workflow deployment, or independent medical decision-making. The findings require expert review, larger-scale external validation, and prospective clinical evaluation before any clinical use.

## Funding

The author received funding for this work.

## Competing Interests

The author declares no competing interests.

## Author Contributions

W.M. designed the study, implemented the model and evaluation pipeline, conducted the experiments, analyzed the results, and wrote the manuscript.

## Acknowledgements

The author expresses great thanks to Victoria Wendy Hughes (King George Secondary School) for thoughtful mentorship and for support in shaping the proposal and the broader scientific direction of the project. The author also thanks Bridge Education and Andrew Lin for academic guidance, technical support, and constructive discussions throughout the development of this study, and thanks David Duan for collaborative discussion during the project.

## Conference Note

A shorter version of this work has been accepted for presentation at the 2026 3rd International Conference on Artificial Intelligence and Future Education (AIFE 2026), Tokyo, Japan. The present manuscript reports an expanded computational imaging preprint version of the work.